\documentclass[conference]{IEEEtran}

\usepackage{amssymb,amsfonts,amsmath,amsthm,amscd,dsfont,mathrsfs}
\usepackage{graphicx,float,psfrag,epsfig,subfigure}
\usepackage{wrapfig}

\DeclareMathAlphabet{\mathpzc}{OT1}{pzc}{m}{it}

\newtheorem{propo}{Proposition}[section]
\newtheorem{lemma}[propo]{Lemma}

\newtheorem{coro}[propo]{Corollary}
\newtheorem{thm}[propo]{Theorem}

\def\cH{{\cal H}}

\def\eps{\epsilon}

\def\reals{{\mathds R}}
\def\naturals{{\mathds N}}

\def\Mmax{M_{\rm max}}

\def\E{\mathbb E}

\def\<{\langle}
\def\>{\rangle}

\def\cP{{\cal P}}
\def\cF{{\cal F}}
\def\chF{\widehat{\cal F}}

\def\eps{\epsilon}

\def\hM{\widehat{M}}

\def\hZ{\widehat{Z}}

\def\E{{\mathbb E}}

\def\hN{\widehat{N}}

\def\bx{\beta_x}
\def\by{\beta_y}
\def\hbx{\hat{\beta}_x}
\def\hby{\hat{\beta}_y}

\def\tX{\widetilde{X}}
\def\tY{\widetilde{Y}}
\def\cH{{\cal H}}
\def\tz{\tilde{z}}

\def\bU{{\overline{U}}}
\def\bV{{\overline{V}}}

\def\Xp{X_{\perp}}
\def\Yp{Y_{\perp}}
\def\de{{\rm d}}

\def\optspace{{\sc OptSpace}}

\begin{document}

\title{Regularization for Matrix Completion}

%\author{Raghunandan~H.~Keshavan\thanks{Department of  Electrical Engineering, Stanford University}\;\;\; and 
 %       Andrea~Montanari${}^{*,}$\thanks{Departments of Statistics, Stanford University}}

\author{\IEEEauthorblockN{Raghunandan~H.~Keshavan\IEEEauthorrefmark{1}
\;\; and\;\;  Andrea Montanari\IEEEauthorrefmark{1}\IEEEauthorrefmark{2}}
\IEEEauthorblockA{Departments of Electrical Engineering\IEEEauthorrefmark{1}
 and Statistics\IEEEauthorrefmark{2}, Stanford University}
%\IEEEauthorblockA{Email: \{raghuram, montanari\}@stanford.edu}
\vspace{-0.5cm}
}

% make the title area
\maketitle
%
%============================================================
%
\begin{abstract}
We consider the problem of reconstructing a low rank
matrix from noisy observations of a subset of its entries.
This task has applications in statistical learning, computer vision, 
and signal processing. 
In these contexts, `noise' generically refers to any contribution to the data 
that is not captured by the low-rank model.
In most applications, the noise level is large compared to the underlying
signal and it is important to avoid overfitting. In order
to tackle this problem, we define a \emph{regularized} cost 
function well suited for spectral reconstruction methods.
Within a random noise model, and in the large system limit,
we prove that the resulting accuracy undergoes
a phase transition depending on the noise level and on the fraction
of observed entries. 
The  cost function can be minimized using \optspace\,
(a manifold gradient descent algorithm).
Numerical simulations show that this approach is 
competitive with state-of-the-art alternatives.
\end{abstract}
%
%============================================================
%

\vspace{-0.075cm}

\section{Introduction}

Let $N$ be an $m\times n$ matrix which is `approximately' low rank, that is
\begin{eqnarray}
N=  M + W = U\Sigma V^T +W\, .\label{eq:MatrixForm}
\end{eqnarray}
where $U$ has dimensions $m \times r$, 
$V$ has dimensions $n\times r$, and $\Sigma$ is a diagonal 
$r\times r$ matrix. Thus $M$ has rank $r$ and $W$ 
can be thought of as noise, or `unexplained contributions' to $N$.
Throughout the paper we assume the normalization $U^TU = m\, I_{r\times r}$
and $V^TV = n\,I_{r\times r}$ ($I_{d\times d}$ being the $d\times d$ identity).

Out of the $m\times n$ entries of $N$, a subset 
$E\subseteq[m]\times [n]$ is observed.
We let  $\cP_E(N)$ be the $m\times n$ matrix that contains the
observed entries of $N$, and is
filled with $0$'s in the other positions
 \begin{align}
  \cP_E(N)_{ij} = \left\{
              \begin{array}{rl}
              N_{ij} & \text{if } (i,j)\in E\, ,\\
              0       & \text{otherwise.}
              \end{array} \right. \label{eq:RevealedMatrixForm}
  \end{align}
The \emph{noisy matrix completion} problem requires to reconstruct
the low rank matrix $M$ from the observations $\cP_E(N)$.
In the following we will also write $N^E = \cP_E(N)$ for the 
sparsified matrix. 
Over the last year, matrix completion has attracted significant
attention because of its relevance --among other applications--
to colaborative filtering. In this case, the matrix $N$ contains
evaluations of a group of customers on a group of products, and one is
interested in exploiting a sparsely filled matrix to provide 
personalized recommendations \cite{Net06}.

In such applications,
the noise $W$ is not a small perturbation and it is 
crucial to avoid overfitting. For instance,
in the  limit $M\to 0$,  the estimate of $\hM$ risks to be
a low-rank approximation of the noise $W$, which would be grossly incorrect.

In order to overcome this problem, we propose
in this paper an algorithm based on minimizing the following 
cost function
\begin{align}
\cF_E(X,Y;S) \equiv 
\frac{1}{2}||\cP_E(N-XSY^T)||_F^2+\frac{1}{2}\lambda\,||S||_F^2\, . \label{eq:mainfunction}
\end{align}
Here the minimization variables are $S\in\reals^{r\times r}$,
and $X \in \reals^{m \times r}$, $Y \in \reals^{n\times r}$ 
with $X^TX = Y^TY = I_{r\times r}$. Finally, $\lambda>0$ is a regularization
parameter.
% with $\lambda$ large corresponding to more
%constrained fit.
%
%*************************************
%

\vspace{-0.1cm}

\subsection{Algorithm and main results}

The algorithm is an adaptation of the \optspace\, algorithm
developed in \cite{KOM09}. A key observation is that the
following modified cost function can be minimized by singular value 
decomposition (see Section \ref{thm:Main}):
\begin{align}
\chF_E(X,Y;S) \equiv 
\frac{1}{2}||\cP_E(N)-XSY^T||_F^2+\frac{1}{2}\lambda\,||S||_F^2\, .
\end{align}
As emphasized in \cite{KOM09,KOM09noisy}, which analyzed the case 
$\lambda=0$, this minimization can yield poor results unless
the set of observations $E$ is `well balanced'. This problem 
can be bypassed by `trimming' the set $E$, and constructing a balanced set
$\widetilde{E}$. The \optspace\, algorithm is given as follows.

\vspace{0.3cm}

\begin{tabular}{ll}
\hline
\vspace{-.35cm}\\
\multicolumn{2}{l}{ {\sc OptSpace} ( set $E$, matrix $N^E$ )}\\
\hline
\vspace{-.35cm}\\
1:& Trim $E$, and let $\widetilde{E}$ be the output;\\
2:& Minimize $\chF_{\widetilde{E}}(X,Y;S)$ via SVD,\\ 
& let $X_0,Y_0,S_0$ be the output;\\
3:& Minimize 
$\cF_E(X,Y;S)$ by gradient descent \\
& using $X_0,Y_0,S_0$ as initial condition.\\
\hline
\end{tabular}

\vspace{0.3cm}

In this paper we will study this algorithm under a model for which
step 1 (trimming) is never called, i.e. $\widetilde{E}=E$
with high probability. We will therefore not discuss it
any further. Section \ref{sec:Simulation} compares the behavior
of the present approach with alternative schemes.
Our main analytical result is a sharp characterization of the mean square
error after step 2. Here and below the limit $n\to\infty$
is understood to be taken with $m/n\to\alpha\in (0,\infty)$.
\begin{thm}\label{thm:Main}
Assume $|M_{ij}|\le M_{\rm max}$, $W_{ij}$ to be i.i.d. 
random variables with mean $0$ variance $\sqrt{mn}\sigma^2$ and 
$\E\{W_{ij}^4\}\le Cn^2$, 
and that for each entry $(i,j)$, $N_{ij}$ is observed (i.e. 
$(i,j)\in E$) independently with probability $p$. 
Finally let $\hM = X_0S_0Y_0^T$ be the rank $r$ matrix reconstructed by step
$2$ of \optspace, for the optimal choice of $\lambda$. 
Then, almost surely for $n\to\infty$
\begin{eqnarray*}
&\hspace{-1cm}&\frac{1}{||M||_F^2}||\hM-M||_F^2 =  1 -  \\
&\hspace{-1cm}&-\frac{\Big\{\sum_{k=1}^r\Sigma_k^2\Big(1-\frac{\sigma^4}{p^2\Sigma^4_k}\Big)_+ 
\Big\}^2}
{
%\Big\{\sum_{k=1}^r\Sigma_k^2\Big\} 
||\Sigma||_F^2
\Big\{\sum_{k=1}^r\Sigma_k^2\Big(1+
\frac{\sqrt{\alpha}\sigma^2}{p\Sigma_k^2}\Big)
\Big(1+\frac{\sigma^2}{p\Sigma_k^2\sqrt{\alpha}}\Big)\Big\}}+o_n(1)\, .
\end{eqnarray*}
\end{thm}
This theorem focuses on a high-noise regime, and predicts 
a sharp phase transition:
if $\sigma^2/p< \Sigma_1$, we can successfully extract information on 
$M$, from the observations $N^E$. If on the other hand
$\sigma^2/p\ge \Sigma_1$, the observations are essentialy useless in 
reconstructing $M$. It is possible to prove \cite{KMfuture}
that the resulting tradeoff between noise and observed entries is tight:
no algorithm can obtain relative mean square error
smaller than one for $\sigma^2/p\ge \Sigma_1$, under a simple random model 
for $M$. To the best of our knowledge, this is the first sharp phase 
transition result for low rank matrix completion.

For the proof of Theorem \ref{thm:Main}, we refer to Section
\ref{sec:Proof}.
An important byproduct of the proof is that it provides a rule
for choosing the regularization parameter $\lambda$, in the large system limit.
%
%*************************************
%

\vspace{-0.cm}

\subsection{Related work}

The importance of regularization in matrix completion is well known to
practitioners. For instance, one important component of  many 
algorithms competing for the Netflix challenge
\cite{Net06}, consisted in minimizing the 
cost function
$\cH_E(X,Y;S)  \equiv 
\frac{1}{2}||\cP_E(N-\tX\tY^T)||_F^2+
\frac{1}{2}\lambda\,||\tX||_F^2
 + \frac{1}{2}\lambda\,||\tY||_F^2$
(this is also known as \emph{maximum margin matrix factorization}
\cite{Srebro1,Srebro2}).
Here the minimization variables are $\tX\in\reals^{m\times r}$,
$\tY\in\reals^{n\times r}$. Unlike in \optspace, these matrices are not
constrained to be orthogonal, and as a consequence the problem becomes
significantly more degenerate. Notice that, in our approach, the orthogonality
constraint fixes the norms $||X||_F$, $||Y||_F$. This motivates the use
of $||S||_F^2$ as a regularization term.

Convex relaxations of the matrix completion problem
were recently studied in \cite{CaR08,CandesPlan}. As emphasized
by Mazumder, Hastie and Tibshirani \cite{HastieEtAl},
such nuclear norms relaxations can be viewed as spectral regularizations
of a least square problem. Finally, the phase transition 
phenomenon in Theorem \ref{thm:Main}, generalizes a result
of Johnstone and Lu on principal component analysis 
\cite{SparsePCA}, and similar random matrix models were studied in 
\cite{Capitaine}.

%
%*****************************************************************
%
\section{Numerical simulations}
\label{sec:Simulation}

In this section, we present the results of numerical simulations 
on synthetically generated matrices. The data are generated 
following the recipe of \cite{HastieEtAl}: sample 
$\bU \in \reals^{n\times r}$ and $\bV \in \reals^{m\times r}$ by choosing 
$\bU_{ij}$ and $\bV_{ij}$ 
independently and indentically as ${\cal N}(0,1)$. Sample independently 
$W \in \reals^{m \times n}$ by choosing $W_{ij}$ iid 
with distribution ${\cal N}(0,\sigma^2\sqrt{mn})$. Set $N = \bU\bV^T + W$.
We also use the parameters chosen in \cite{HastieEtAl} and
define 
\begin{eqnarray*}
{\rm SNR} & = & \sqrt{\frac{{\rm Var}((\bU\bV^T)_{ij})}{{\rm Var}(W_{ij})}}\, ,  \\
{\rm Test Error} & = & \frac{||\cP_{ E}^{\perp}(\bU\bV^T - \widehat{N})||_F^2 }{  ||\cP_{ E}^{\perp} (UV^T)||_F^2   } \, ,\\
{\rm Train Error} & = & \frac{||\cP_{ E }(N - \widehat{N})||_F^2 }{  ||\cP_{ E }(N)||_F^2   }\, ,
\end{eqnarray*}
where $\cP_{ E}^{\perp}(A) \equiv A-\cP_E(A)$.

\begin{figure}
\begin{center}
\subfigure{
\includegraphics[width=8cm]{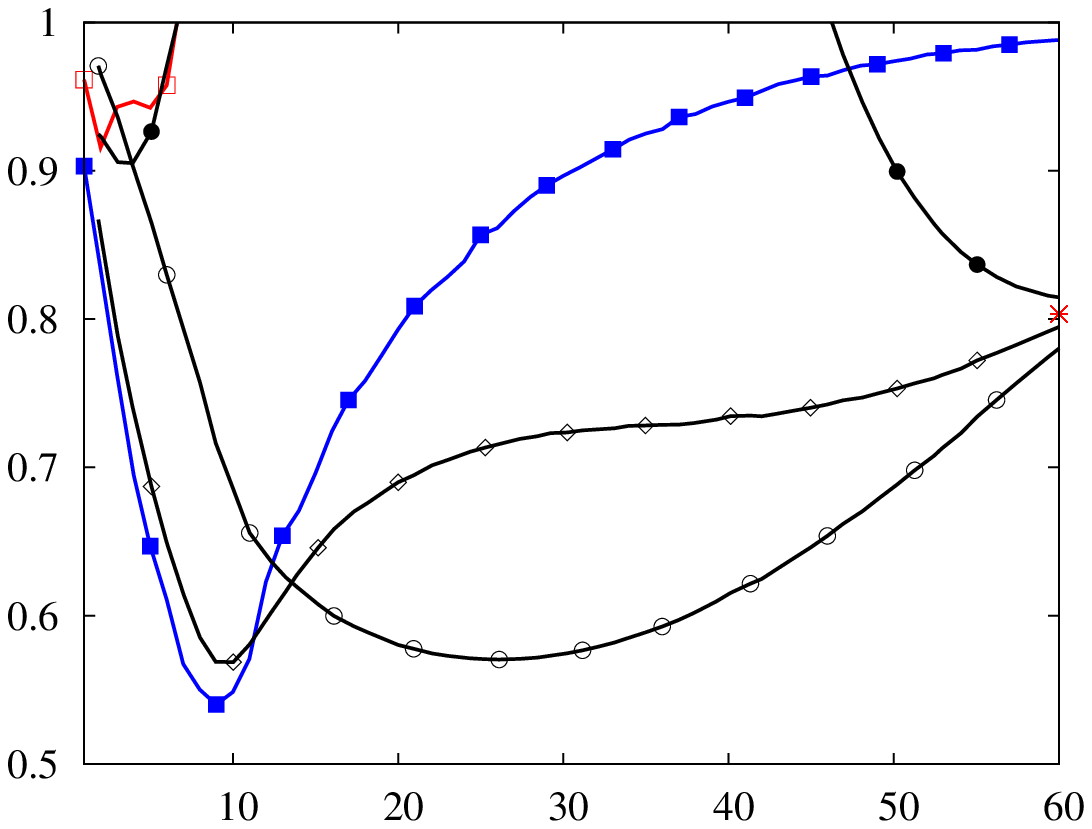}
}
\subfigure{
\includegraphics[width=8cm]{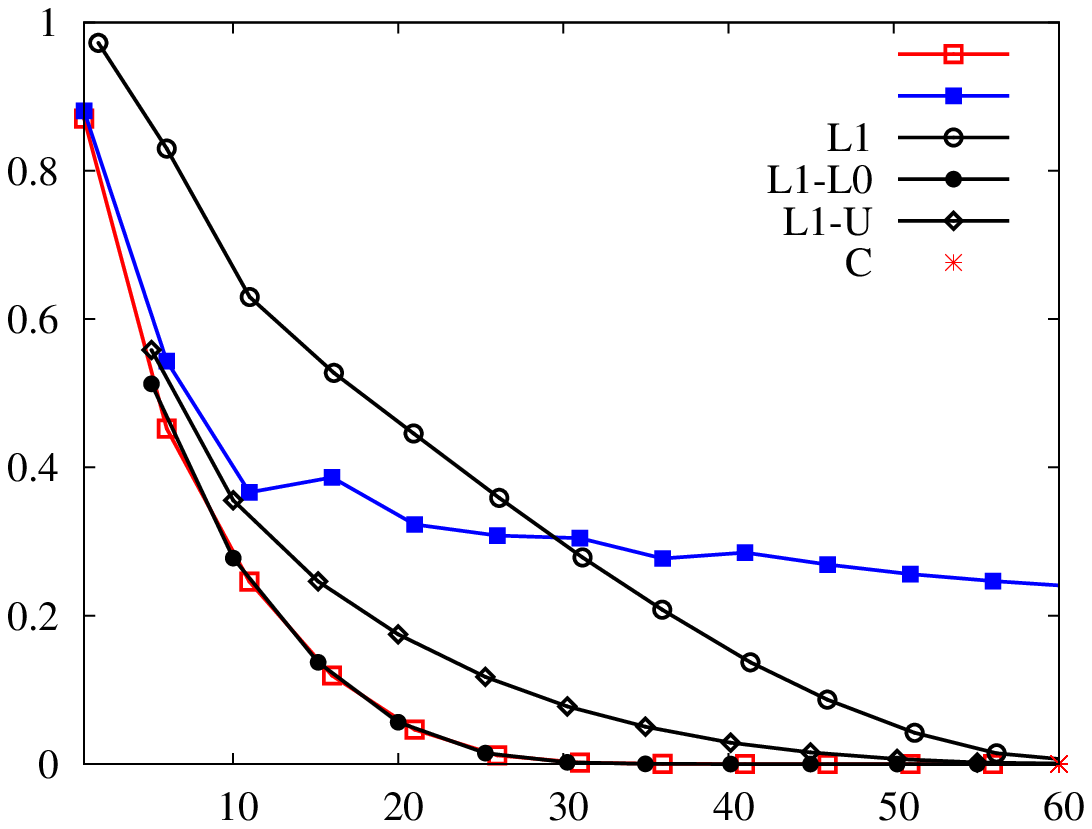}
\put(-98,135){ {\footnotesize {\sc OptSpace}}{\footnotesize ($\lambda^*$)} }
\put(-98,142){ {\footnotesize {\sc OptSpace}}{\footnotesize ($0$)} }
}
\caption{Test (top) and train (bottom) error vs. rank for \optspace, {\sc Soft-Impute}, {\sc Hard-Impute} and SVT. Here $m=n=100,r=10,p = 0.5,{\rm SNR}=1$. } \label{fig:r10}
\end{center}
\end{figure}

\begin{figure}
\begin{center}
\subfigure{
\includegraphics[width=8cm]{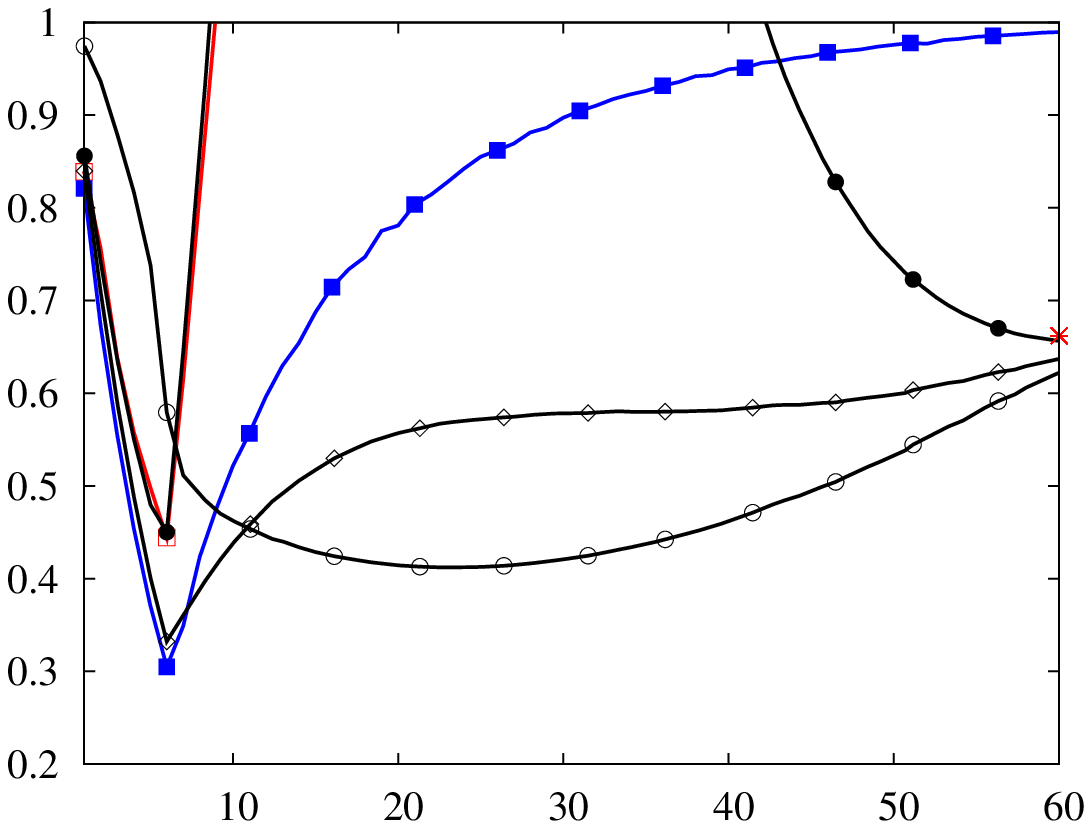}
}
\subfigure{
\includegraphics[width=8cm]{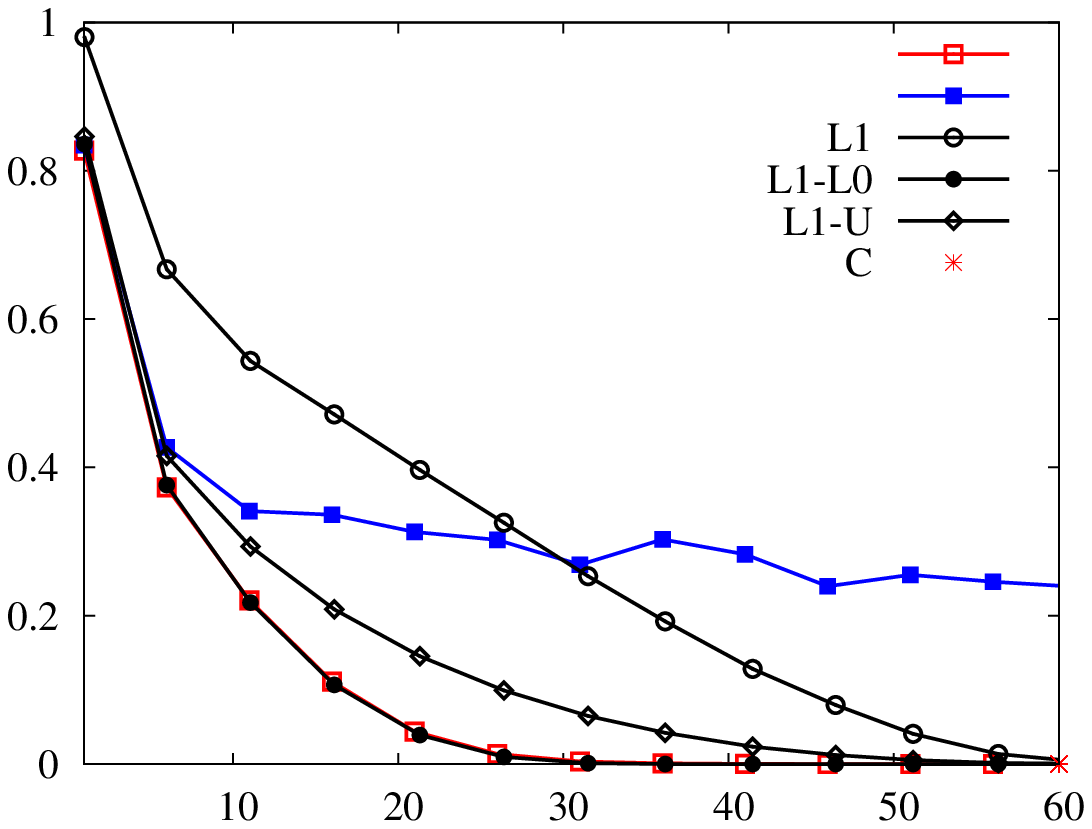}
\put(-98,135){ {\footnotesize {\sc OptSpace}}{\footnotesize ($\lambda^*$)} }
\put(-98,142){ {\footnotesize {\sc OptSpace}}{\footnotesize ($0$)} }
}
\caption{Test (top) and train (bottom) error vs. rank for \optspace, {\sc Soft-Impute}, {\sc Hard-Impute} and SVT. Here $m=n=100,r=6, p = 0.5,{\rm SNR}=1$. } \label{fig:r6}
\end{center}
\end{figure}

\begin{figure}
\begin{center}
\subfigure{
\includegraphics[width=8cm]{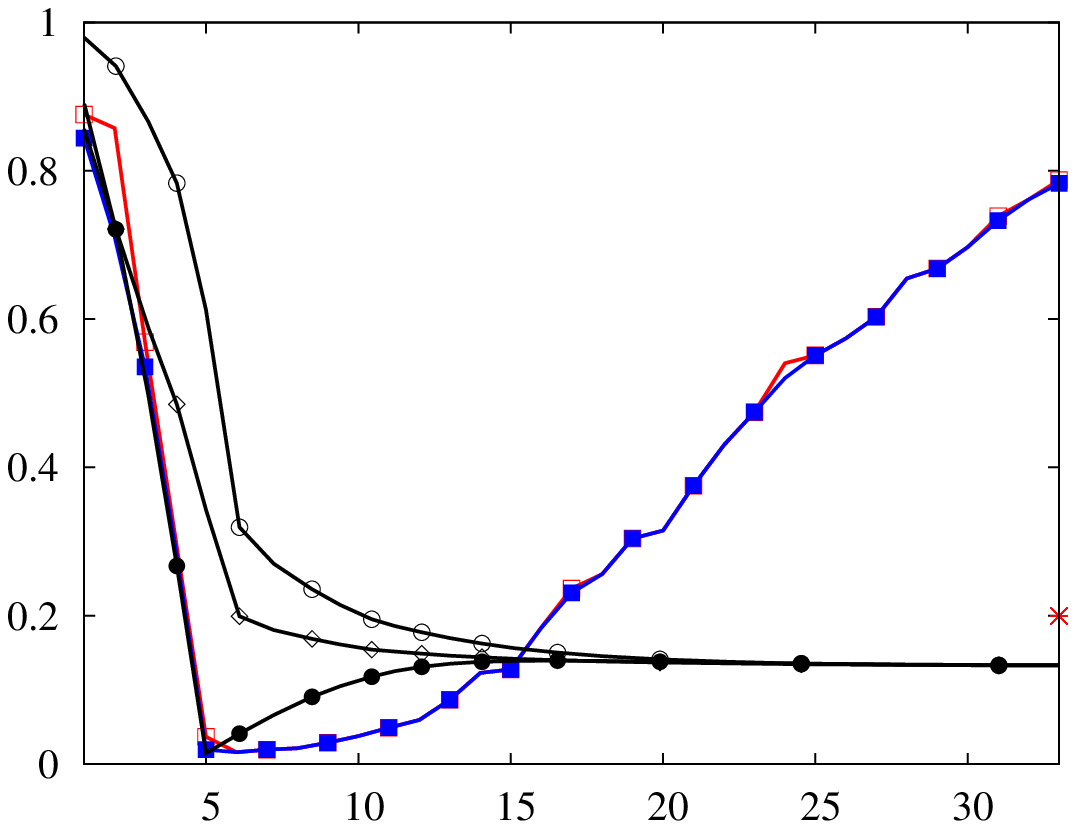}
}
\subfigure{
\includegraphics[width=8cm]{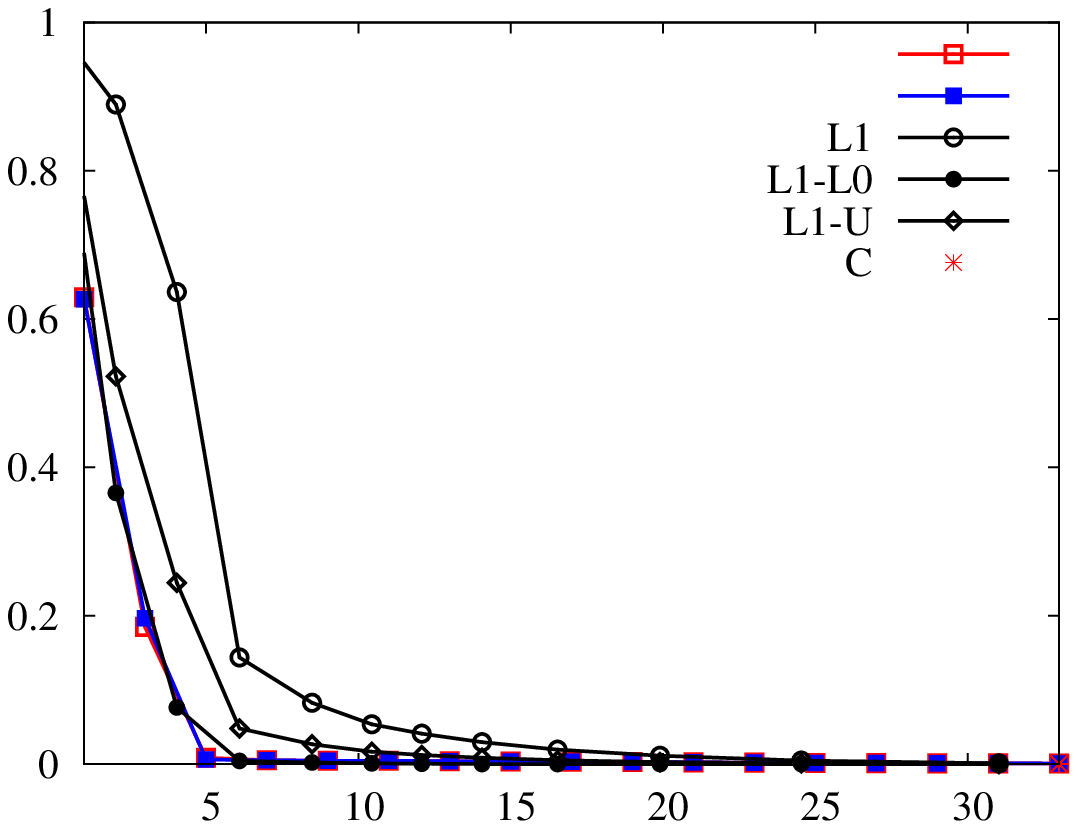}
\put(-98,135){ {\footnotesize {\sc OptSpace}}{\footnotesize ($\lambda^*$)} }
\put(-98,142){ {\footnotesize {\sc OptSpace}}{\footnotesize ($0$)} }
}
\caption{Test (top) and train (bottom) error vs. rank for \optspace, {\sc Soft-Impute}, {\sc Hard-Impute} and SVT. Here $m=n=100,r=5, p = 0.2,{\rm SNR}=10$. } \label{fig:r5}
\end{center}
\end{figure}

In Figure \ref{fig:r10}, we plot the train error and test error for the 
{\sc OptSpace} algorithm on matrices generated as above with $n=100, r=10$, 
SNR=$1$ and $p = 0.5$. For comparison, we also plot the corresponding 
curves for {\sc Soft-Impute},{\sc Hard-Impute} and SVT taken from 
\cite{HastieEtAl}.  In Figures \ref{fig:r6} and \ref{fig:r5}, we plot 
the same curves for different values of $r,\eps,{\rm SNR}$. In these 
plots, \optspace$(\lambda)$ corresponds to the algorithm that minimizes 
the cost (\ref{eq:mainfunction}). In particular \optspace$(0)$ corresponds 
to the algorithm described in \cite{KOM09}. Further, 
$\lambda^* = \lambda^*(\rho)$ is the value of the regularization 
parameter that minimizes the test error while using rank $\rho$
(this can be estimated on a subset of the data, not used for training).  

It is clear that regularization greatly improves the performance
of \optspace\, and makes it competitive with the best alternative methods.

%
%*****************************************************************
%
\section{Proof of Theorem \ref{thm:Main}}
\label{sec:Proof}

The proof of Theorem 1 is based on the following three steps:
$(i)$ Obtain an explicit expression for the root mean square error
in terms of right and left singular vectors of $N$;
$(ii)$ Estimate the effect of the noise $W$ on the
right and left singular vectors; $(iii)$ Estimate the effect of
missing entries. Step $(ii)$ builds on recent estimates on the 
eigenvectors of large covariance matrices 
\cite{BaiMiaoPan}. In step $(iii)$ 
we use the results of \cite{KOM09}.
Step $(i)$ is based on the following linear algebra calculation,
whose proof we omit due to space constraints
(here and below $\<A,B\>\equiv {\rm Tr}(AB^T)$).
\begin{propo}
Let $X_0\in\reals^{m\times r}$ and $Y_0\in\reals^{m\times r}$
be the matrices whose columns are the first $r$,
right and left, singular vectors of $N^E$.
Then the rank-$r$ matrix reconstructed by step
$2$ of of \optspace, with regularization parameter $\lambda$,
has the form $\hM(\lambda) = X_0S_0(\lambda)Y_0^T$
Further, there exists $\lambda_*>0$ such that 
\begin{align}
\frac{1}{mn}
||M-\hM(\lambda_*)||_{F}^2 = 
||\Sigma||_F^2 - \left( \frac{ \<X_0^TMY_0 \, , \,X_0^T N^E Y_0\> }{\sqrt{mn}||X_0 N^E Y_0||_F}\right)^2 \, .\label{eq:GeneralExpression}
\end{align}
\end{propo}
%
%
%************************************************
%
\subsection{The effect of noise}

In order to isolate the effect of noise, we consider the 
matrix $\hN =  p\, U \Sigma V^T + W^E$. Throughout this section
we assume that the hypotheses of Theorem \ref{thm:Main} hold.
\begin{lemma}\label{lemma:Noise}
Let $(n z_{1,n},\dots , nz_{r,n})$ be the $r$ largest 
singular values of $\hN$. Then, as $n\to\infty$, $z_{i,n}\to z_i$
almost surely, where, for $\Sigma_i^2> \sigma^2/p$,
\begin{eqnarray}
z_i =  p \Sigma_i
\left\{ \alpha \left( \frac{\sigma^2}{p\Sigma_i^2} + \frac{1}{\sqrt{\alpha}} \right) \left( \frac{\sigma^2}{p \Sigma_i^2} + \sqrt{\alpha} \right) \right\}^{1/2}\, ,\label{eq:Zis}
\end{eqnarray}
and $z_i =  \sigma \sqrt{p\alpha^{1/2}}(1+\sqrt{\alpha})$ 
for $\Sigma_i^2\le \sigma^2/p$.

Further, let  $X \in \reals^{m \times r}$ and $Y\in \reals^{n \times r}$ 
be the matrices whose columns are the first $r$,
right and left, singular vectors of $\hN$.
Then there exists a sequence of $r\times r$ orthogonal 
matrices $Q_n$ such that, almost surely
$||\frac{1}{\sqrt{m}}U^TX - A Q_n||_F \to 0$,
$||\frac{1}{\sqrt{n}}V^TY - B Q_n||_F \to 0$
with $A = {\rm diag}(a_1,\dots, a_r)$, $B = {\rm diag}(b_1,\dots, b_r)$
and 
\begin{eqnarray}
a_i^2 & = & \Big(1 - \frac{\sigma^4}{p^2 \Sigma_i^4}\Big)
\Big(1 + \frac{\sqrt{\alpha} \sigma^2}{p \Sigma_i^2}  \Big)^{-1}\, , \nonumber \\
%\;\;\;\;\;\;\;\;\;
b_i^2 & = & \Big( 1 - \frac{\sigma^4}{p^2 \Sigma_i^4}\Big)
\Big( 1 + \frac{\sigma^2}{p\sqrt{\alpha}  \Sigma_i^2}  \Big)^{-1}\, ,
\end{eqnarray}
for $\Sigma_i^2> \sigma^2/p$, while $a_i = b_i=0$ otherwise.
\end{lemma}
\begin{IEEEproof}
Due to space limitations, we will focus here on the case 
$\Sigma_1,\dots,\Sigma_r>\sigma^2/p$. The general proof proceeds
along the same lines, and we defer it to \cite{KMfuture}.

Notice that $W^E$ is an $m\times n$ matrix with i.i.d. entries with 
variance $\sqrt{mn}\sigma^2p$ and fourth moment bounded by $Cn^2$.
It is therefore sufficient to prove our claim for $p=1$ and then rescale 
$\Sigma$ by $p$ and $\sigma$ by $\sqrt{p}$. We will also assume that,
without loss of generality, $m\ge n$. 

Let $\hZ$ be an $r\times r$ diagonal matrix containing the eigenvalues
$(nz_{n,1},\dots,nz_{n,r})$. The eigenvalue equations read
\begin{eqnarray}
U \hby + WY - X\hZ & = & 0\, ,  \label{eq:main1}\\
V \hbx + W^TX - Y\hZ & = & 0\, . \label{eq:main2}
\end{eqnarray}
where we defined
$\hbx \equiv \Sigma\, U^TX $, $\hby \equiv \Sigma\,
 V^TY\in\reals^{r\times r}$.
By singular value decomposition we can write
$W = L \, {\rm diag}(w_1, w_2, \ldots w_n) R^T$, with $L^TL = I_{m\times m}$, 
$R^TR=I_{n\times n}$. 

Let $u_i^T$, $x_i^T$, $v_i^T$, $y_i^T\in\reals^r$ 
be the $i$-th row of -respectively-  $L^TU$, $L^TX$, $R^TV$, $R^TY$.
In this basis equations ($\ref{eq:main1}$) and ($\ref{eq:main2}$)
read
\begin{eqnarray*}
u_i^T \hby + w_iy_i^T - x_i^T\hZ& = & 0\, ,   \qquad i \in [n]\, , \\
u_i^T \hby - x_i^T\hZ & = & 0\, , \qquad i \in [m] \backslash [n]\, , \\
v_i^T\hbx + w_ix_i^T - y_i^T\hZ & = & 0\, ,  \qquad i\in [n]\, .
\end{eqnarray*}
These can be solved to get
\begin{eqnarray}
x_i^T & = & (u_i^T\hby \hZ + w_i v_i^T \hbx)(Z^2 - w_i^2)^{-1}\, ,  
\qquad i \in [n]\, , \nonumber \\
x_i^T & = & u_i^T \hby \hZ^{-1}\, ,  \qquad\qquad
\qquad\qquad\qquad\;\;\;\;\;\; i \in [m] \backslash [n]\, , \nonumber\\
y_i^T & = & (v_i^T \hbx \hZ +  w_i u_i^T \hby )(\hZ^2 - w_i^2)^{-1}\, , \qquad 
i \in [n].\label{eq:SolYi}
\end{eqnarray}
By definition $\Sigma^{-1}\hbx = \sum_{i=1}^m u_ix_i^T$,
and $\Sigma^{-1}\hby = \sum_{i=1}^n v_iy_i^T$, whence
\begin{eqnarray}
\Sigma^{-1} \hbx & \hspace{-.4cm}  = \hspace{-.4cm} &\sum_{i=1}^{n} u_i(u_i^T \hby \hZ + w_i v_i^T \hbx)(\hZ^2 - w_i^2)^{-1} \nonumber \\
& & + \sum_{i=n+1}^{m} u_iu_i^T \by \hZ^{-1} \, ,
\label{eq:LinearAlgebra1}\\
\Sigma^{-1} \hby & \hspace{-.4cm} = \hspace{-.4cm} & \sum_{i=1}^{n} v_i(v_i^T \hbx \hZ + w_i u_i^T \hby)(\hZ^2 - w_i^2)^{-1}.\label{eq:LinearAlgebra2}
\end{eqnarray}

Let $\lambda = w_i^2\alpha^{1/2}/(m^2\sigma^2)$. Then, it is a well known 
fact \cite{SilvBai95}
that as $n\to\infty$ the empirical law of the $\lambda_i$'s
converges weakly almost surely to the Marcenko-Pastur law, with density
$\rho(\lambda) = \alpha\sqrt{(\lambda-c_-^2)(c_+^2-\lambda)}/(2\pi\lambda)$,
with $c_{\pm} = 1\pm \alpha^{-1/2}$.

Let $\bx = \hbx/\sqrt{m}$, $\by=\hbx/\sqrt{n}$, $Z = \hZ/n$.
A priori, it is not clear that the sequence $(\bx,\by,Z)$ --dependent on 
$n$-- converges. However, it is immediate to show that the sequence 
is tight, and hence we can restrict ourselves to a subsequence
$\Xi\equiv \{n_i\}_{i\in\naturals}$ along which a 
limit exists.
Eventually we will show that the limit does not depend on the subsequence,
apart, possibly, from the rotation $Q_n$. Hence we shall denote the
subsequential limit, by an abuse of notation, as $(\bx,\by,Z)$.

Consider now a such a convergent subsequence. It is possible to
show that $\Sigma_i^2> \sigma^2/p$ implies 
$Z_{ii}^2>\alpha^{3/2} \sigma^2 c_+(\alpha)^2+\delta$ for some positive 
$\delta$. Since almost surely as $n\to\infty$, $w_i^2<\alpha^{3/2} 
\sigma^2 c_+(\alpha)^2+\delta/2$ for all $i$, for all purposes the summands 
on the rhs of Eqs.~(\ref{eq:LinearAlgebra1}), (\ref{eq:LinearAlgebra2}) 
can be replaced by uniformly continuous, bounded functions
of the limiting eigenvalues $\lambda_i$. Further, each entry
of $u_i$ (resp. $v_i$) is just a single coordinate of the left (right)
singular vectors of the random matrix $W$. Using 
Theorem $1$ in \cite{BaiMiaoPan}, it  follows that
any subsequential limit satisfies the equations
\begin{align}
\bx &  =  \Sigma\by\Big\{ 
Z \int ( Z^2 - \alpha^{3/2} \sigma^2 \lambda)^{-1} 
\rho(\lambda) 
\de\lambda + (\alpha - 1) Z^{-1}\Big\} \, , \label{eq:bxbyeq1}\\
\by &  =  \Sigma \bx \Big\{ 
Z \int ( Z^2 -  \alpha^{3/2} \sigma^2 \lambda)^{-1}\, \rho(\lambda)
\, \de\lambda \Big\}, .\label{eq:bxbyeq2}
\end{align}
 
Solving for $\by$, we get an equation of the form
\begin{eqnarray}
\Sigma^{-2} \by  = \by  \, f(Z) \label{eq:betaequation}
\end{eqnarray}
where $f(\,\cdot\,)$ is a function that can be given 
explicitely using the Stieltjis transform of the measure 
$\rho(\lambda)\de\lambda$.
Equation (\ref{eq:betaequation}) implies that $\by$ is block diagonal 
according to the degeneracy pattern of $\Sigma$.
Considering each block, either $\beta_y$ vanishes in the block
(a case that can be excluded using $\Sigma_i^2>\sigma^2/p$)
or $\Sigma_i^{-2} = f(Z_{ii})$ in the block. Solving
for $Z_{ii}$ shows that the eigenvalues are uniquely determined
(independent of the subsequence) and given by Eq.~(\ref{eq:Zis}).

In order to determine $\bx$ and $\by$ first observe
that, since $I_{r\times r}= Y^TY =\sum_{i=1}^ny_iy_i^T$, we have,
using Eq.~(\ref{eq:SolYi})
\begin{eqnarray*}
I_{r\times r}  & = &
\sum_{i=1}^{n} (\hZ^2 - w_i^2)^{-1}  ( \hZ \hbx^T v_i +  w_i \hby^T u_i )  \\
& & (v_i^T \hbx \hZ +  w_i u_i^T \hby )(\hZ^2- w_i^2)^{-1}\,.
\end{eqnarray*}
In the limit $n\to\infty$, and assuming a convergent subsequence
for $(Z,\bx,\by)$, this sum can be computed as above. After
\begin{eqnarray*}
I_{r\times r} & = & \Big\{ \int \frac{Z^2}{(Z^2 - \alpha^{3/2}\sigma^2 
\lambda)^2} \rho(\lambda)\,\de\lambda \Big\} C_x \\
& & + \Big\{ \int \frac{\alpha^{3/2}\sigma^2 \lambda}{(Z^2 - \alpha^{3/2} 
\sigma^2\lambda)^2} \rho(\lambda)\,\de\lambda  \Big\} C_y \, ,
\end{eqnarray*} 
where $C_x = \bx^T\bx$, $C_y = \by^T\by$ and 
the functions of $Z$  on the rhs are defined as standard analyic 
functions of matrices.

Using  Eqs.~(\ref{eq:bxbyeq1}), (\ref{eq:bxbyeq2})
and solving the above, we get
$C_x = {\rm diag}(\Sigma_1^2 a_1^2, \dots \Sigma_r^2 a_r^2)$,
and 
$B_y = {\rm diag}(\Sigma_1^2 b_1^2, \dots \Sigma_r^2 b_r^2)$.
We already concluded that $\bx$ and $\by$ are block diagonals with 
blocks in correspondence with the degeneracy  pattern of 
$\Sigma$. Since  $\bx^T\bx=C_x$ and $\by^T\by=C_y$ are diagonal,
with the same degeneracy pattern, it follows that, inside each block
of size $d$, each of 
$\bx$ and $\by$ is proportional to a $d\times d$ orthogonal matrix.
Therefore $\bx = \Sigma A Q_s$, $\by = \Sigma B Q'_s$,
for some othogonal matriced $Q_s$, $Q_s'$. 
Also, using equation (\ref{eq:bxbyeq1}) one can prove
that  $Q_s=Q_s'$.

Notice, by the above argument $A$, $B$ are uniquely
fixed by our construction. On the other hand $Q_s$ might 
depend on the subsequence $\Xi$. Since our statmement allows
for a seqence of rotations $Q_n$, that depend on $n$,
the eventual subsequence dependence of $Q_s$ can be factored out.
\end{IEEEproof}

It is useful to point out a straightforward consequence of the above.
\begin{coro}
There exists a sequence of orthogonal matrices $Q_n\in\reals^{r\times r}$
such that, almost surely,
\begin{eqnarray}
\lim_{n\to\infty}\Big|\Big|\frac{1}{\sqrt{mn}}
X^TU\Sigma V^T Y-Q_n D Q^T_n\Big|\Big|_F & = & 0\, , 
\label{eq:finalbxby}
\end{eqnarray}
with $D = {\rm diag}(\Sigma_1a_1b_1,\dots,\Sigma_ra_rb_r)$.
\end{coro}

%
%********************************************************
%
\subsection{The effect of missing entries}

The proof of Theorem \ref{thm:Main} is completed by 
estabilishing a relation between the singular vectors $X_0$, $Y_0$ of 
$N^E$ and the singular vectors $X$ and $Y$ of $\hN$. 
\begin{lemma}\label{lemma:BoundPerp}
Let $k\le r$ be the largest integer 
such that $\Sigma_1\ge\dots \ge\Sigma_k>\sigma^2/p$,
and denote by $X_0^{(k)}$, $Y_0^{(k)}$, $X^{(k)}$, and $Y^{(k)}$ the matrices
containing the first $k$ columns of
$X_0$, $Y_0$, $X$, and $Y$, respectively.
Let $X^{(k)}_0 = X^{(k)} S_x + X^{(k)}_{\perp}$,
$Y^{(k)}_0 = Y^{(k)}S_y + Y^{(k)}_{\perp}$ where 
$(X^{(k)}_{\perp})^TX^{(k)} = 0$, $(Y^{(k)}_{\perp})^TY^{(k)} = 0$ and 
$S_x,S_y\in\reals^{r\times r}$. Then there exists a numerical constant 
$C = C(\Sigma_i, \sigma^2,\alpha,M_{\rm max})$,
such that, with high probability,
\begin{eqnarray}
||\Xp^{(k)}||_F^2, ||\Yp^{(k)}||_F^2 \le C r\,  \sqrt{ \frac{1}{n} } \,
,
\end{eqnarray}
with probability approaching $1$ as $n\to\infty$.
\end{lemma}
\begin{IEEEproof}
We will prove our claim for the right singular vector $Y$, since the
left case is completely analogous. Further we will drop the superscript
$k$ to lighten the notation.

We start by noticing that $||N^{E} Y_0||_F^2=\sum_{a=1}^k (n\tz_{a,n})^2$,
where $n\tz_{a,n}$ are the singular values of $N^E$. Using  
Lemma 3.2 in \cite{KOM09} which bounds
$||M^E-pM||_2 = ||N^E-\hN||_2$, we get 
\begin{eqnarray} 
||N^{E} Y_0||_F^2& \ge & \sum_{a=1}^{k} \left( n z_{a,n}  
- C \Mmax \sqrt{pn} \right)^2 \, .\label{eq:N1lb}
\end{eqnarray} 

On the other hand 
$||N^E Y_0||_F \le ||\hN Y_0||_F + ||N^E - \hN||_2 ||Y_0||_F$.
Further by letting $S_y=L_y\Theta_yR_y^T$, for $L_y, R_y$ orthogonal
matrices, we get
$||\hN Y_0||_F^2  =  ||\hN YL_y\Theta_y||_F^2 + ||\hN \Yp||_F^2$.
Since $Y_0^TY_0 = I_{k\times k}$, we have 
$I_{k\times k} = R_y \Theta_y^T \Theta_y R_y^T + Y_{\perp}^T Y_{\perp}$,
and therefore
\begin{eqnarray*}
||\hN Y_0||_F^2
& = & ||\hN Y L_y ||_F^2 - || \hN Y L_y R_y^T\Yp^T ||_F^2 +  ||\hN \Yp||_F^2 \\
& \le & n^2 \sum_{a=1}^{k} z_{a,n}^2 - n^2 z_{k,n}^2 ||\Yp||_F^2 \\ 
& &+ n^2 p \sigma^2 \alpha (c_+(\alpha)+\delta)||\Yp||_F^2 \\
& = & n^2 \sum_{a=1}^{k} z_{a,n}^2 - n^2 e_y ||\Yp||_F^2\, ,
\end{eqnarray*}
where $e_y \equiv z_{k,n}^2 - p\sigma^2 \alpha (c_+(\alpha)+\delta)$,
and used the inequality $||\hN \Yp||_F^2\le n^2 p \sigma^2 \alpha 
(c_+(\alpha)+\delta)||\Yp||_F^2 $ which holds for all $\delta>0$ 
asymptotically almost surely as $n\to\infty$
(by an immediate generalization of
Lemma \ref{lemma:Noise}).
It is simple to check that $\Sigma_k \ge \sigma^2 /p$ 
implies $e_y>0$. 

Using triangular inequality, Lemma 3.2 in \cite{KOM09}, we get
\begin{eqnarray*}
||N Y_0||_F^2 & \le &  n^2 \sum_{a=1}^{r} z_{a,n}^2 - n^2 e_y ||\Yp||_F^2 + C np \alpha^{3/2} \Mmax^2 r\\
& & + 2C n\sqrt{np} \alpha^{3/4} \Mmax \sqrt{r} ||z|| \, ,
\end{eqnarray*}
which, combined with equation (\ref{eq:N1lb}), implies the thesis.
\end{IEEEproof}

\begin{IEEEproof}[Proof of Theorem \ref{thm:Main}]
We now turn to upper bounding the right hand side of
Eq.~(\ref{eq:GeneralExpression}). Let $k$ be defined as in the last 
lemma.  
Notice that by Lemma \ref{lemma:Noise}, $X^T(U\Sigma V^T)Y$ is well
approximated by $(X^{(k)})^T(U\Sigma V^T)Y^{(k)}$. 
Analogously, it can be proved that $X^T_0(U\Sigma V^T)Y_0$ is well
approximated by $(X^{(k)}_0)^T(U\Sigma V^T)Y^{(k)}_0$.
Due to space limitations, we will omit this technical step
and thus focus here on the case $k =r$ (equivalently, neglect
the error incurred by this approximation).

Using Lemma \ref{lemma:BoundPerp} to bound the contribution
of $X_{\perp},Y_{\perp}$, we have 
\begin{eqnarray}
&\hspace{-1.5cm}&\< X_0^T(U\Sigma V^T)Y_0 \, , \,  X_0^T N^E Y_0\> \nonumber \\
&\hspace{-1cm}& = \< S_x^T X^T(U\Sigma V^T)Y S_y \, , \,  X_0^T N^E Y_0\> (1+  o_n(1)) \nonumber\\
&\hspace{-1cm}& = \< X^T(U\Sigma V^T)Y \, , \,  S_x^T X_0^T N^E Y_0 S_y \> (1+ o_n(1))
\label{eq:BeforeLast} \, .
\end{eqnarray}
Further  $X_0^T N^E Y_0= X_0^T \hN Y_0+X_0^T (N^E-\hN)Y_0$ and,
using once more the bound in Lemma 3.2 of \cite{KOM09},
that implies $|X_0^T (N^E-\hN)Y_0|\le Cr\sqrt{nrp}$, we get
\begin{eqnarray*}
S_x^T X_0^T N^E Y_0 S_y  & = &
L_x \Theta_x^2 L_x^T X^T \hN Y R_y \Theta_y^2  R^T_y + E_1 \\
& = & Z + E_2 \, ,
\end{eqnarray*}
where we recall that $Z$ is the diagonal matrix with entries 
given by the singular values of $\hN$, and 
$||E_1||_F^2, ||E_2||_F^2 \le C(p,r) \sqrt{n}$.
Using this estimate in Eq.~(\ref{eq:BeforeLast}), together
with the result in Lemma \ref{lemma:Noise}, we finally get
\begin{eqnarray*}
\frac{ \< X_0^T (U\Sigma V^T)Y_0 \, , \,  X_0^T N^E Y_0\> }{\sqrt{mn} || X_0^T N^E Y_0 ||_F^2 }\ge \frac{\sum_{k=1}^{r}  \Sigma_k a_k b_k z_k }{\sqrt{\alpha} ||z||}  - o_n(1)\, ,
\end{eqnarray*}
which implies the thesis after simple algebraic manipulations
\end{IEEEproof}

%*****************************************************************
\section*{Acknowledgements}

We are grateful to T.~Hastie, R.~Mazumder and R.~Tibshirani for 
stimulating discussions, and for making available their data.
This work was supported by
a Terman fellowship, and the NSF grants CCF-0743978
and  DMS-0806211.

%
%*****************************************************************
%
\bibliographystyle{IEEEtran}

\bibliography{MatrixCompletion}

% Generated by IEEEtran.bst, version: 1.13 (2008/09/30)
\begin{thebibliography}{10}
\providecommand{\url}[1]{#1}
\csname url@samestyle\endcsname
\providecommand{\newblock}{\relax}
\providecommand{\bibinfo}[2]{#2}
\providecommand{\BIBentrySTDinterwordspacing}{\spaceskip=0pt\relax}
\providecommand{\BIBentryALTinterwordstretchfactor}{4}
\providecommand{\BIBentryALTinterwordspacing}{\spaceskip=\fontdimen2\font plus
\BIBentryALTinterwordstretchfactor\fontdimen3\font minus
  \fontdimen4\font\relax}
\providecommand{\BIBforeignlanguage}[2]{{%
\expandafter\ifx\csname l@#1\endcsname\relax
\typeout{** WARNING: IEEEtran.bst: No hyphenation pattern has been}%
\typeout{** loaded for the language `#1'. Using the pattern for}%
\typeout{** the default language instead.}%
\else
\language=\csname l@#1\endcsname
\fi
#2}}
\providecommand{\BIBdecl}{\relax}
\BIBdecl

\bibitem{Net06}
``Netflix prize,'' {\tt http://www.netflixprize.com/}.

\bibitem{KOM09}
R.~H. Keshavan, A.~Montanari, and S.~Oh, ``Matrix completion from a few
  entries,'' January 2009, {\tt arxiv:0901.3150}.

\bibitem{KOM09noisy}
------, ``Matrix completion from noisy entries,'' June 2009, arXiv:0906.2027.

\bibitem{KMfuture}
R.~H. Keshavan and A.~Montanari, ``Regularization for matrix completion,''
  2010, journal version, in preparation.

\bibitem{Srebro1}
N.~Srebro, J.~Rennie, and T.~Jaakkola, ``Maximum margin matrix factorization,''
  in \emph{Advances in Neural Information Processing Systems 17}, 2005.

\bibitem{Srebro2}
J.~Rennie and N.~Srebro, ``Fast maximum margin matrix factorization for
  collaborative prediction,'' in \emph{22nd International Conference on Machine
  Learning}, 2005.

\bibitem{CaR08}
E.~J. Cand{\`e}s and B.~Recht, ``Exact matrix completion via convex
  optimization,'' \emph{Found. of Comput. Math.}, vol.~9, no.~6, pp. 717 --
  772, 2009.

\bibitem{CandesPlan}
E.~J. Cand{\`e}s and Y.~Plan, ``Matrix completion with noise,'' 2009, {\tt
  arXiv:0903.3131}.

\bibitem{HastieEtAl}
R.~Mazumder, T.~Hastie, and R.~Tibshirani, ``Spectral regularization algorithms
  for learning large incomplete matrices,'' 2009, submitted.

\bibitem{SparsePCA}
I.~M. Johnstone and A.~Y. Lu, ``On consistency and sparsity for principal
  component analysis in high dimension,'' \emph{J. Amer. Stat. Assoc.}, vol.
  104, pp. 682--693, 2009.

\bibitem{Capitaine}
M.~Capitaine, C.~Donati-Martin, and D.~F\'eral, ``The largest eigenvalue of
  finite rank deformation of large wigner matrices: convergence and
  non-universality of the fluctuations,'' \emph{Ann. Probab.}, vol.~37, pp.
  1--47, 2009.

\bibitem{BaiMiaoPan}
Z.D.Bai, B.Q.Miao, and G.M.Pan, ``On asymptotics of eigenvectors of large
  sample covariance matrices,'' \emph{Ann. of Probab.}, vol.~35, pp.
  1532--1572, 2007.

\bibitem{SilvBai95}
J.~Silverstein and Z.~Bai, ``On the empirical distribution of eigenvalues of a
  class of large-dimentional random matrices,'' \emph{J. Multivariate Anal.},
  vol.~54, pp. 175--192, 1995.

\end{thebibliography}

\end{document}